\pdfoutput=1
\documentclass[11pt]{article}

\usepackage[preprint]{acl}
\usepackage{times}
\usepackage{latexsym}

\usepackage[T1]{fontenc}
\usepackage[utf8]{inputenc}

\usepackage{microtype}

\usepackage{inconsolata}

\usepackage{graphicx}
\usepackage{todonotes}
\usepackage{booktabs}
\usepackage{verbatim}
\usepackage{colortbl} % For cell background colors
\usepackage{xcolor} % For custom colors
\usepackage{booktabs} % For professional table lines
\usepackage{array} % For column alignment
\usepackage{tcolorbox}
\usepackage{framed}
\usepackage{enumitem}
\usepackage{setspace}
\usepackage{listings}
\usepackage{xurl}
\usepackage{hyperref}
\usepackage{array}

\title{Effective Performance Measurement:  \\ Challenges and Opportunities in KPI Extraction from Earnings Calls} 

\author{
\textbf{Rasmus T.  Aavang\textsuperscript{1,2}},
  \textbf{Rasmus Tjalk-Bøggild\textsuperscript{2}},
  \textbf{Alexandre Iolov\textsuperscript{2}},
  \textbf{Giovanni Rizzi\textsuperscript{2}},
  \\
  \textbf{Mike Zhang\textsuperscript{3}},
  \textbf{Johannes Bjerva\textsuperscript{1}}
\\
\\
  \textsuperscript{1}Department of Computer Science, Aalborg University, Denmark \\
  \textsuperscript{2}ALIPES ApS, Denmark \\
  \textsuperscript{3}University of Copenhagen
\\
  \small{
      \textbf{Correspondence:} \href{mailto:rtaj@cs.aau.dk}{rtaj@cs.aau.dk}
  }
}
\usepackage{multirow}
\usepackage{float}
\usepackage{subcaption}
\usepackage{adjustbox}
\usepackage{graphicx}
\usepackage{booktabs} % For better table lines
\usepackage{fancyvrb} % For verbatim environments
\usepackage{courier}  % For the Courier font
\usepackage{makecell}  % For the Courier font
\usepackage{tabularx}

\newcommand{\modelname}[1]{{#1}}

\begin{document}
\maketitle
\begin{abstract}
Earnings calls are a key source of financial information about public companies. 
However, extracting information from these calls is difficult.
Unlike the templatic filings required by the U.S. Securities and Exchange Commission (SEC) to report a company's financial situation, earnings conference calls have no built-in labels, are unstructured, and feature conversational language.
We explore this challenging domain by assessing the information captured by models trained on SEC filings and in-context learning methods. 
To establish a baseline, we first evaluate the generalization capabilities of SEC-trained models across established SEC datasets.
To support our investigation, we introduce three novel benchmarks: (1) SEC Filings Benchmark (\textbf{SECB}), (2) Earnings Calls Benchmark (\textbf{ECB}), and  \textbf{ECB-A}, a subset with 2,460 expert annotation groups to support our qualitative analysis.
We find that encoder-based models struggle with the domain shift. 
Finally, we propose a system utilizing LLMs to perform open-ended extraction from unstructured call transcripts, verified by human evaluation (79.7\% precision), providing a baseline for this valuable domain through the consistent tracking of emergent KPIs.
\end{abstract}

\definecolor{Blue}{HTML}{1F77B4} % Blue
\definecolor{Green}{HTML}{2CA02C} % Green
\definecolor{Orange}{HTML}{FF7F0E} % Orange
\definecolor{Red}{HTML}{D62728} % Red

\section{Introduction}
The efficient market hypothesis~\cite{fama1970efficient} states that, given current prices already incorporate all public information, the primary driver of significant price change is new information.
Consequently, industry investors are in a constant \emph{race for new information}, as a company's valuation can swing 40\% in seconds (Figure~\ref{fig:frontpage}).
Two of the most critical public disclosures in this race are SEC filings and corresponding earnings conference calls. 
Earnings calls comprise two key components: (i) Senior management's presentation of \emph{key performance indicators} (KPIs) and (ii) Q\&A session with analysts and investors. 
As a result, calls play a crucial role for industry investors in assessing a company's value.
However, extracting information from earnings calls is difficult; neither gold annotations nor a method for automatic KPI extraction exists.
We address key challenges in the emerging task of information extraction from earnings calls by evaluating both state-of-the-art methods, developed initially for SEC filings, and in-context learning as illustrated in Figure~\ref{fig:page1}. 
Our guiding RQs:

\begin{figure}[t]
    \centering
    \includegraphics[width=\linewidth, trim=0 0cm 0 0, clip]{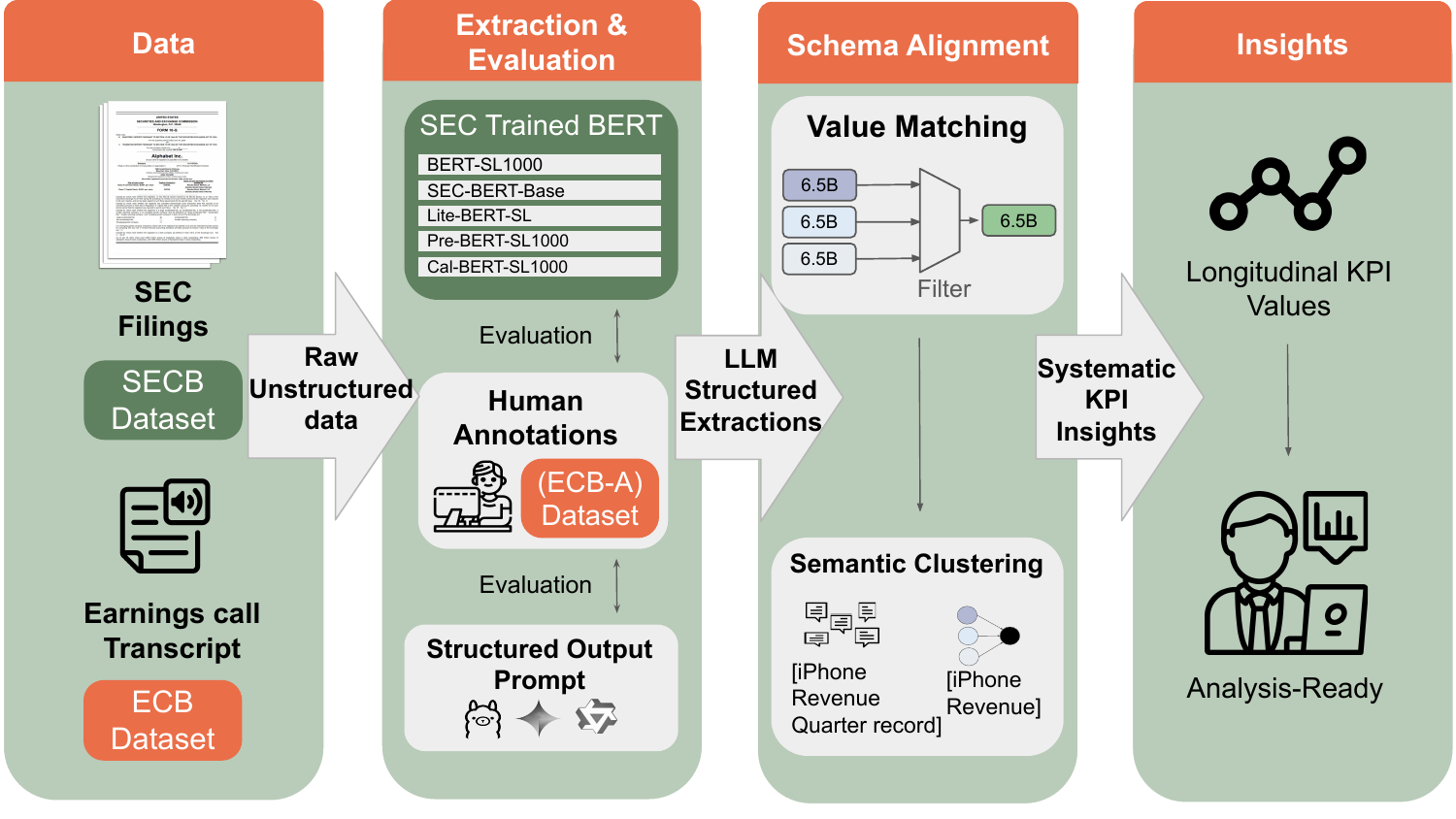}
    \caption{\textbf{Analysis and Pipeline.} 
    We ground our analysis in the established SEC filings domain. 
    To capture the open-ended set of KPIs in earnings calls, we adopt a relation extraction strategy to benchmark encoders and in-context learning against expert annotations. 
    Finally, we aggregate structured outputs to generate consistent, longitudinal KPI tracking suitable for financial analysis.}
    \label{fig:page1}
\end{figure}

\textbf{RQ1} How well do encoder-based models, fine-tuned on structured SEC filings, generalize across the linguistic shift to the unstructured domain of earnings conference calls?

\textbf{RQ2} How well can current State-of-the-Art (SOTA) Large Language Models perform structured KPI extraction, and what actionable impact can be derived from the current performance?

\section{Background}
\paragraph{Automated KPI Extraction}
Financial NLP has shifted from encoders like FinBERT \citep{araci2019finbert, yang2020finbert} to focus on larger models such as FinMA \citep{xie2023pixiulargelanguagemodel} and InvestLM \citep{yang2023investlmlargelanguagemodel}, proprietary tools like Bloomberg GPT~\citep{wu2023bloomberggpt}, and RAG-based FinGPT~\citep{yang2023fingptopensourcefinanciallarge}.
Despite this progress, work targeting KPI extraction is limited and usually relies on pre-defined schemas.
KPI-BERT~\citep{hillebrand2022kpibertjointnamedentity} tries to link values to descriptions in German financial documents.
FiNER-139 \citep{loukas-etal-2022-finer} and HiFi-KPI \citep{aavang2025hifi} classify entities into fixed taxonomies. %(e.g., 139 labels or hierarchical definitions derived from standardized and templatic SEC Filings).
Less investigated is KPI extraction from earnings calls, despite its impact on investment returns~\citep{chen2018manager, qin-yang-2019-say, ma2020towards, barahona2024impact}, likely because of the free-flowing, unstructured, and unlabeled nature of the calls.
To address this unlabeled nature, we adopt a relation extraction approach~\citep{etzioni2008open}.
Inspired by recent advancements in open-domain extraction like ODKE+ \citep{khorshidi2025odkeontologyguidedopendomainknowledge}, we utilize state-of-the-art LLMs: Llama-3.3, Qwen3-30B-A3B,
Gemma-3-27B-it and Gemini 3 pro~\citep{grattafiori2024llama3herdmodels, yang2025qwen3technicalreport, gemmateam2025gemma3technicalreport, gemini3_2025} to dynamically extract KPIs without prior schemas.

\paragraph{Financial Theory of Performance Measurement}
Various methods of performance measurement have been explored -- see \citet{khan2011understanding} for an overview. 
We adopt the definitions of \citet{ghalayini1996changing}, distinguishing between \emph{traditional} (retrospective, financial) and \emph{non-traditional} (operational, forward-looking) metrics (Appendix Table~\ref{tab:Tradional}).
While prior work focuses on the extraction of traditional KPIs. 
We also consider non-traditional KPIs, more present in earnings calls and deemed crucial in the assessment of business performance \citep{ghalayini1996changing}.

\paragraph{SEC Filings} 
Publicly traded American companies follow the regulations of the Securities and Exchange Commission (SEC) and file 10-Qs (Quarterly updates) and 10-Ks (Annual updates). \citep{sec_exchange_2024}.
10-Qs and 10-Ks are highly templatic and focus on traditional performance measures.

\paragraph{Earnings Calls}

\definecolor{earnings_report}{HTML}{5f815f} 
\definecolor{earnings_call}{HTML}{ea6e48} 
\definecolor{Q_A}{HTML}{C8D8C8}
\begin{figure}[t]
    \centering
\includegraphics[width=\linewidth, trim={4.5cm 0 4.5cm 0}, clip]{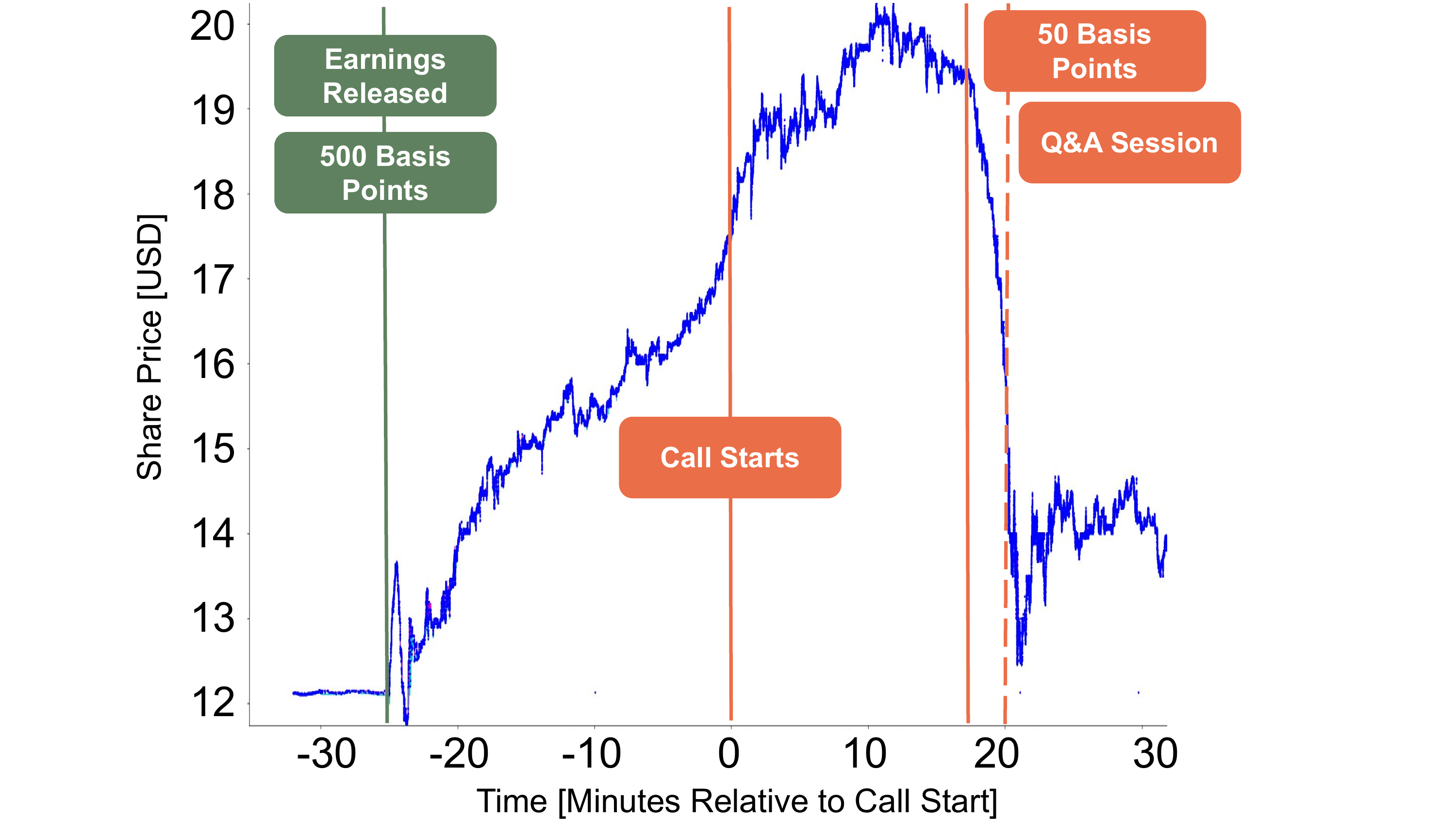}
    \caption{\textbf{Lyft's share price} from the release of its earnings report to the end of the earnings call.  
    When the incorrect value is presented in the \textcolor{Green}{earnings release} the price rises quickly. 
    However, once the error is corrected during the \textcolor{earnings_call}{earnings call}, the price rapidly drops.}
    \label{fig:frontpage}
\end{figure}

During these calls, management and investors discuss financial results \citep{investopedia_earnings_call}. 
Figure \ref{fig:frontpage} shows the race for new information in practice.
An error in the reported value for the KPI "EBITDA margin expansion"  increased the company's valuation by over \$3 billion, before a correction in the call prompted a \$3 billion drop.
Conference calls are not legally mandated, but most companies do them with the structure:
\begin{enumerate}
          
    \item \textbf{Presentation and Discussion 
    }\\[0.5em]
          Management, usually the CFO and CEO, discuss the financial results and present KPIs \citep{cfi_earnings_call}.
          
    \item \textbf{Q\&A Session}\\[0.5em]
          Usually, earnings calls end with a Q\&A session, where investors and analysts -- and today even retail investors -- can vote on potential questions to ask \citep{markov2023democratization}.
\end{enumerate}

\paragraph{LM-Based Metrics} 
Language model-based metrics like BERTScore \citep{zhang2019bertscore} and cross-encoders (e.g., STSB-RoBERTa-large) surpass rule-based methods \citep{reimers2019sentence, ebrahim2024warwicknlp}. 
LLMs demonstrate robust judgment capabilities \citep{zheng2023judging}, though prone to self-preference bias \citep{wataoka2024self}. 
Consequently, we employ DeepSeek-V3.2 \citep{deepseekai2025deepseekv32pushingfrontieropen} as a distinct evaluator, complemented by a RoBERTa-based score.

\section{Experimental Setup}

\begin{table}[t]
\centering
\resizebox{\columnwidth}{!}{
\begin{tabular}{lccc}
\toprule
& \textbf{\# Entries} & \textbf{Period} &   \textbf{Entities} \\
\midrule
\textbf{FiNER-139} & 1.1M & 2016-2020 &  387K \\
\textbf{HiFi-KPI} (Lite) & 1.9M(8.0K) & 2017-06/2024 &  5,300K \\
\textbf{SECB} & 41K & 2023-2024 &  78K \\
\textbf{ECB} (ECB-A) & 10.5K (587) & 2023-2024 &  (2.5K) \\
% there are 619 chunks 86,426 words.
\bottomrule
\end{tabular}
}
\caption{\textbf{Dataset Statistics:} Comparison of earlier datasets and our benchmarks (\textbf{SECB}, \textbf{ECB}, \textbf{ECB-A}).}
\label{tab:merged_dataset_statistics}
\end{table}
We construct three novel datasets from 20 S\&P 500 companies\footnote{The selected tickers are: \texttt{AAPL}, \texttt{JNJ}, \texttt{JPM}, \texttt{AMZN}, \texttt{BA}, \texttt{PG}, \texttt{XOM}, \texttt{NEE}, \texttt{GOOGL}, \texttt{DOW}, \texttt{PLD}, \texttt{MSFT}, \texttt{PFE}, \texttt{BAC}, \texttt{HD}, \texttt{CAT}, \texttt{KO}, \texttt{CVX}, \texttt{DUK}, and \texttt{SHW}.}.
We obtain SEC filings from \citet{sec_website} and earnings call transcripts from \citet{financialmodelingprep}.
{ 
\addtolength{\leftmargini}{-6pt}
\begin{enumerate}
    \setlength{\itemsep}{-6pt} 
    \item {SEC Filings Benchmark (\textbf{SECB})}
    \item {Earnings Call Benchmark (\textbf{ECB})}
    \item {Earnings Call Benchmark Annotated (\textbf{ECB-A})} 
\end{enumerate}
}
\noindent Table \ref{tab:merged_dataset_statistics} compares these to existing resources also included in our analysis. 
To bridge the gap between templated filings and conversational calls, we optimize \textbf{SECB} to preserve broader context, than HiFi-KPI~\citep{aavang2025hifi}, which prioritizes KPI density.
Within this expanded ``free text,'' we introduce the pseudo-tags \textbf{\textit{regex dollar}} and \textbf{\textit{regex percentage}} to capture unannotated KPIs.
\textbf{ECB} has 10,477 chunks from 2023--2024 earnings calls across 20 companies, segmented by speaker turns (uninterrupted speech by the same speaker). 
To ensure reliability, we establish \textbf{ECB-A}, an evaluation subset consisting of 10 randomly selected full transcripts. 
A domain expert performed a two-stage annotation process separated by a 3-month break: an initial pass to identify KPI descriptors and values, followed by a confirmation pass to verify entities and connect entity relations. 
\textbf{ECB-A} contains 587 chunks (avg. 147 words) annotated with 2,460 entities (4.19/chunk) and 934 relational groups (1.59/chunk). 
(Annotation and parsing details in \textbf{Appendix \ref{app:detailed_setup} and \ref{ManualAnnoSetup}}).
 
\paragraph{Experiments} 
We investigate KPI extraction from earnings call transcripts using two distinct paradigms: domain-specific encoders (SEC-based BERT models) and in-context learning via few-shot prompting.
We assess the generalization ability across SEC filings-based dataset, by standardizing the labelspace and evaluating \modelname{SEC-BERT-BASE} and \modelname{BERT-SL1000} on \textbf{SECB}, as well as on the {FiNER-139}~\citep{loukas-etal-2022-finer} and HiFi-KPI~\citep{aavang2025hifi} test sets as a sequence labeling task\footnote{Code and datasets are available at \url{https://github.com/aaunlp/effective-performance-measurement}.}. 
Following this baseline analysis, we evaluate performance on unstructured earnings calls using the annotated \textbf{ECB-A} dataset. 
We also test LLMs with a structured few-shot prompting scheme (details in \textbf{Appendix \ref{app:prompt}}). 
This prompt mirrors our expert annotator guidelines: models must first identify KPI entity spans and then aggregate them into relational groups to generate a descriptive label. 
To evaluate the SEC-based encoders against our open-ended relation schema, we map their predicted token-level classes directly to the 'Label' field in our grouping structure.
\paragraph{ECB-A Metrics} 
We base our evaluation on the following metrics with semantic scoring from cross-encoder \textit{STSB-Roberta-Large}~\citep {reimers2019sentence}.
\begin{enumerate}
    \setlength{\itemsep}{-6pt} 
    \item \textbf{Exact F1:} Standard F1 requiring exact string matches for both value and label.
    \item \textbf{Semantic F1:} Derived from the mean maximum similarity of predictions to ground truths (precision) and vice-versa (recall), allowing for many-to-one mapping.
    \item \textbf{Match F1:} A soft F1 score derived from the label similarity of predictions strictly aligned to ground truths by value (precision) and vice-versa (recall), treating unaligned items as zero.
    \item \textbf{LLM Judge:} The percentage of value-grouped ground truths found and evaluated as equivalent by DeepSeek-V3.2
\end{enumerate}
Finally, we utilize \textbf{ECB} to show a practical system employing semantic clustering to identify KPIs.

\section{Empirical Results}
\newcolumntype{C}{>{\centering\arraybackslash}X}
% \section{Experiments}
\begin{table}[t]
\footnotesize
\centering
\begin{tabularx}{\columnwidth}{@{}lCCCC@{}} 
    \toprule
    & \multicolumn{2}{c}{\textbf{µ}-F1} & \multicolumn{2}{c}{\textbf{M}-F1} \\
    \cmidrule(lr){2-3} \cmidrule(lr){4-5}
    Dataset & {SB} & SL1000 & {SB} & SL1000 \\
    \midrule
    \textbf{SECB} & 0.057 & \textbf{0.143} & 0.032 & \textbf{0.192} \\
    FiNER-139 & \textbf{0.842} & 0.662 & \textbf{0.859} & 0.624 \\
    HiFi-KPI & 0.240 & \textbf{0.501} & 0.001 & \textbf{0.012} \\
    \bottomrule
  \end{tabularx}
  \caption{\textbf{SEC-BASED BERT models on SEC Filings Performance.} SEC-BERT-BASE ({SB}) and BERT-SL1000, evaluated on FiNER-139, HiFi-KPI and \textbf{SECB}. Ignoring the "O" label.
 }
  \label{tab:SEC-bert-expanded}
\end{table}

\begin{figure}[t]
    \centering
    \includegraphics[width=\linewidth]{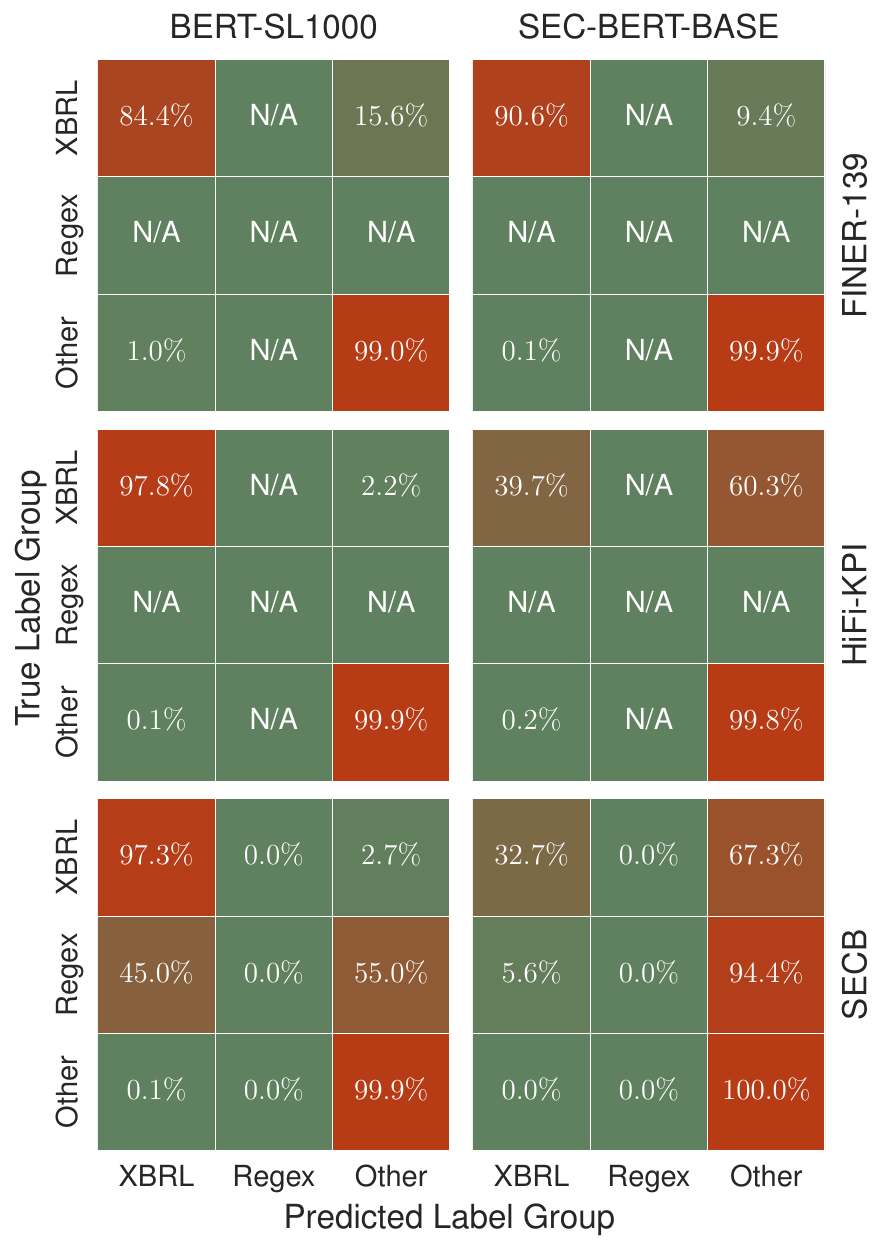}
    \caption{Confusion matrices for \modelname{SEC-BERT-BASE} and \modelname{SL1000} on HiFi-KPI, FiNER-139, and \textbf{SECB}.}
    \label{fig:error_matrices}
\end{figure}

Our experiments reveal a stark contrast in model performance between structured SEC filings and more conversational earnings calls. 
Table \ref{tab:SEC-bert-expanded} shows that while absolute scores vary, neither model experiences a catastrophic drop in performance when shifting datasets.
Although the thousands of labels and regex-based labels in \textbf{SECB} inherently limit absolute performance metrics, the reasonable Micro-F1 scores confirm the models' utility.
In Figure \ref{fig:error_matrices} we simplify gold labels into: \textit{XBRL}, \textit{regex}, and \textit{other}.
We find that \modelname{BERT-SL1000} is significantly more aggressive, classifying 45\% of \textit{Regex} spans as \textit{XBRL}, compared to only 5.6\% for \modelname{SEC-BERT-BASE}. 
This suggests \modelname{BERT-SL1000} may generalize better to the unlabelled financial data found in earnings calls.
However, as shown in Table \ref{tab:model_performance_multilayer}, SEC-trained models fail to generalize to earnings calls; even though \modelname{BERT-SL1000} extracts some correct numbers, it fails to predict compatible labels.
In contrast, the generative models show promise and significantly outperform the SEC-trained models, while exact match performance is still low, the semantic-based metrics show promise for the generative models.
Performance is especially impressive considering the models operate in a fully unconstrained setting, identifying entities directly from text without reliance on a closed taxonomy.

Qwen-3 and Llama-70B find 26.55\% and 33.94\% of the expert annotations according to the LLM judgment, with Gemini 3 pro achieving the best performance with an exact F1 of 11.5\% and extraction of 45.5\% of the annotated KPIs according to the LLM judgment.
While exact F1 is low, semantic F1 is high, especially for Llama-3.3 and Gemini 3 pro.
Despite a high semantic score, Gemma-3's low match rate indicates frequent value-relation misalignment, likely causing its poor LLM-as-a-judge performance.
The significantly higher Semantic and LLM-Judge scores demonstrate that while LLMs capture the underlying financial concepts, they struggle with the strict lexical boundaries of Exact Match extraction.
The continued performance improvement with model scaling, even at the state-of-the-art level, highlights the task's inherent complexity.

\begin{table}[t]
  \centering
  \footnotesize 
      \resizebox{\columnwidth}{!}{%
  \begin{tabular}{lrrrr}
    \toprule    
    & \multicolumn{4}{c}{\textbf{Scores (\%)}} \\
    \cmidrule(lr){2-5}
    \textbf{Model} & \textbf{Exact} & \textbf{Semantic} & \textbf{Match} & \textbf{LLM Judge} \\
    \midrule
    SEC-BERT-BASE & 0.0 & 0.0 & 0.0 & 0.0 \\
    Lite-BERT-SL & 0.0 & 7.1 & 1.6 &  0.0 \\
    Pre-BERT-SL1000 & 0.0 & 5.6 & 1.3 & 0.8 \\ 
    Cal-BERT-SL1000 & 0.0 & 4.8 & 1.5 & 0.9 \\
    BERT-SL1000 & 0.0 & 4.7 & 1.1 & 0.4 \\
    Gemma-3-27B & 3.2 & 40.0 & 11.6 & 8.8 \\
    Qwen3-30B-A3B & 3.5 & 38.1 & 26.2 & 33.9 \\
    Llama-3.3-70B & 3.4 & 51.5 & 25.8 & 26.6 \\
    Gemini 3 Pro   & \textbf{11.5} & \textbf{61.6} & \textbf{39.2} & \textbf{45.5} \\
    \bottomrule
  \end{tabular}}
  \caption{\textbf{Performance Comparison on ECB-A.} }
  \label{tab:model_performance_multilayer}
\end{table}

\section{Industry Application \& ECB}
\begin{figure}[t]
    \centering
    \includegraphics[width=\linewidth]{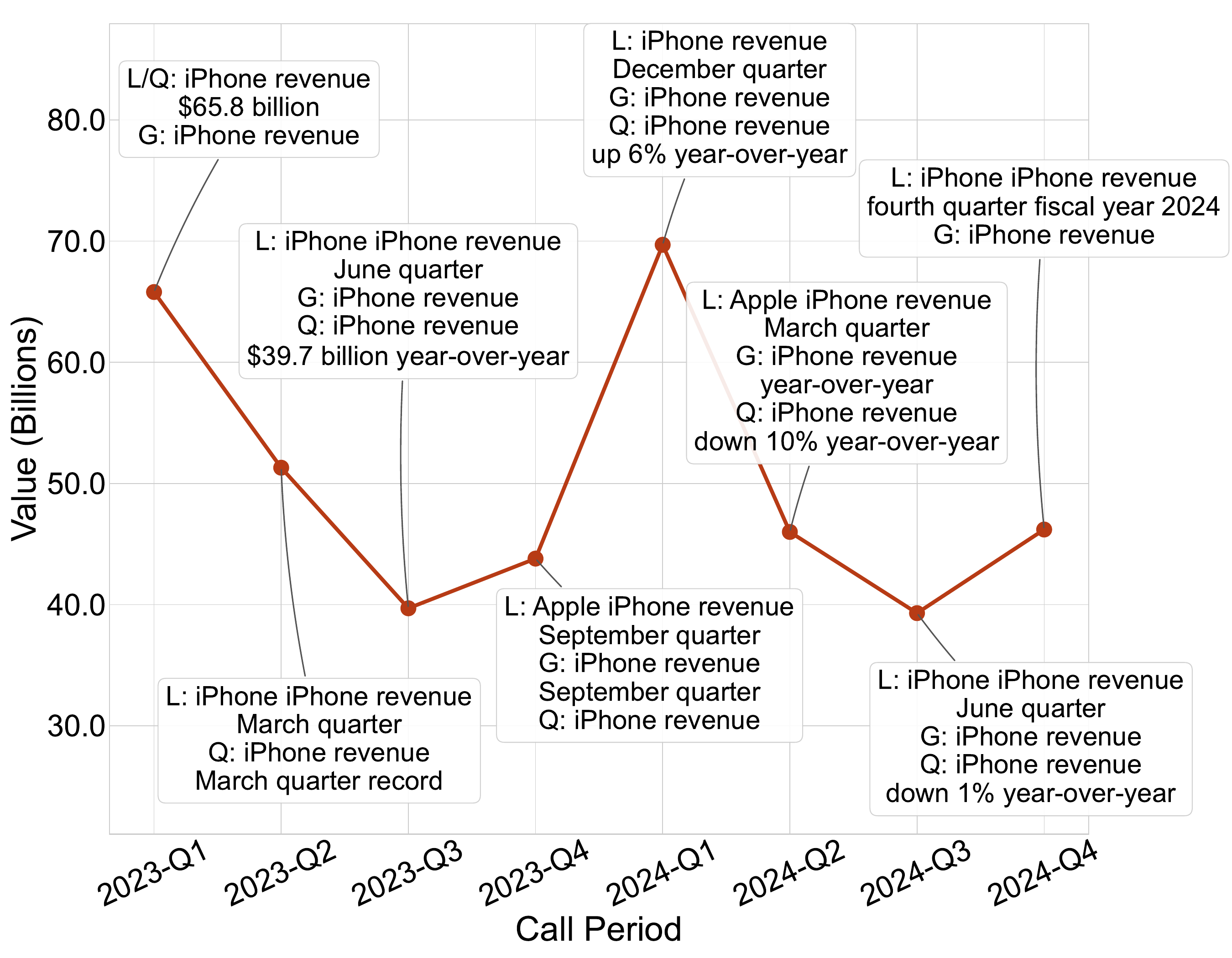} 
    \caption{\textbf{Longitudinal KPI Tracking.} 
    Our system automatically identifies Centroids (e.g., `iPhone revenue'). 
    Callouts display the raw predictions from L (Llama-3.3), G (Gemma-3), and Q (Qwen3).}
    \label{fig:vertical_stack}
\end{figure}

\begin{table}[t]
\centering
\small
% Reduce padding to fit 5 columns in a single column width
\setlength{\tabcolsep}{3.5pt} 
\begin{tabular}{lrrr}
\toprule
\textbf{Model} & \textbf{Share of Pred.} & \textbf{Centroid} & \textbf{Overlap}  \\
 & \textbf{(\%)} & \textbf{(\%)} & \textbf{(\%)} \\
\midrule
Llama-3.3-70B   & 77.63 & 63.19 & 31.26 \\
Qwen3-30B-A3B  & 78.68 & 61.86 & 25.84  \\
Gemma-3-27B   & 77.70 & 59.53 & 30.25 \\
\bottomrule
\end{tabular}
\caption{\textbf{Model Contributions.} All models contribute comparably to the final extraction. 
Llama has the most common centroid, while Qwen has the lowest chance of using the same label as the other models.}
\label{tab:model_agreement}
\end{table}

We scale our experiments to the ECB dataset, comprising two years of longitudinal data across 20 companies.
Figure  \ref{fig:vertical_stack} shows our system enabling KPI discovery, meaning we can consistently discover and track KPIs across time without any defined ontologies.
Since our prompt enforces a strict entity schema, we implement a post-hoc aggregation pipeline mirroring the match metric logic.
Due to compute constraints and reasonable performance, we opt for using \textit{Gemma}, \textit{Llama}, and \textit{Qwen}.
First, we focus on value alignment; we require numerical values to be within a $1\%$ tolerance.
We then grouped extracted KPI entities based on semantic similarity.
We apply a similarity threshold of $0.85$ between all entities' KPI names in the same cluster.
For each identified cluster, we assign a canonical label corresponding to the cluster centroid—defined as the KPI name with the minimum aggregate distance to all other variations in the group.
Finally, we limit the tracking of KPIs to the centroids found in 4 different periods for the same company. 
We default the period to be the current call unless the date entity extracted explicitly mentioned another quarter or year.
Further, we observe a perfect match comparing the values extracted for "iPhone Revenue" by our system with the actual reported values in Apple's SEC filings.
Our system finds 1,323 KPIs that can be consistently tracked across at least 4 periods for the 20 companies in two years of calls.
All 3 models agree on 4.16\% of these KPI extractions. 
Table \ref{tab:model_agreement} shows all models contributing to the final results, even though Gemma-27B showed worse performance in isolation.
\paragraph{Human Evaluation}

\begin{table}[t]
    \centering
    \small
    \begin{tabularx}{\linewidth}{>{\centering\arraybackslash}X >{\centering\arraybackslash}X}
        \toprule
        \textbf{Metric} & \textbf{Score} \\
        \midrule
        Krippendorff's $\alpha$ & 0.429 \\
        Precision & 79.67\% (478/600) \\
        \bottomrule
    \end{tabularx}
    \caption{\textbf{Human Evaluation.} Though Krippendorff’s $\alpha$ (0.429) indicates moderate agreement, the final extraction shows high precision of 79.67\% (478/600).}
    \label{tab:human_eval}
\end{table}

To verify our final extractions via the post-hoc aggregation, we employ three evaluators to verify 200 extractions each (100 overlapping; see Appendix \ref{Evaluation}). 
Evaluators assessed the KPI label and value correctness. 
Krippendorff's $\alpha$~\citep{hayes2007answering} of 0.43 and average Cohen's $\kappa$~\citep{cohen1960coefficient} of 0.39 imply moderate agreement~\cite{wong2021cross}; however, raw agreement remains high (69\%). 
This suggests the low $\alpha$ is dampened by the high positive label prevalence. 
Our high system precision of 79.67\% demonstrates promising extraction capabilities in this domain. 
Which can, however, definitely be improved by future more sophisticated methods.

\section{Analysis}
\begin{table}[t]
    \centering
    \resizebox{\columnwidth}{!}{%
        \begin{tabular}{lrr}
            \toprule
            \textbf{Model} & \textbf{Valid KPI} & \textbf{\# Unmatched} \\
                           & \textbf{(\%)}           & \textbf{Predictions} \\
            \midrule
            Gemma-3-27B-it              & 18\% & 318 \\
            Qwen3-30B-A3B-Instruct-2507 & 18\% & 701 \\
            Llama-3.3-70B-Instruct      & 32\% & 588 \\
            Gemini 3 pro Preview                & 26\% & 771 \\
            \bottomrule
        \end{tabular}%
    }
    \caption{\textbf{Unmatched Predictions.} 
    Llama-3.3 and Gemini 3 Pro yield the highest validity. 
    Llama achieves the peak validity (32\%), likely because of its more conservative extraction volume compared to Gemini.}
    \label{tab:unmatched_predictions}
\end{table}
We begin our analysis with a systematic manual comparison between \textbf{ECB-A} and model predictions. 
We randomly sample 100 unmatched extractions from each model to determine if these discrepancies stem from model error or omissions in the expert annotation.
Table \ref{tab:unmatched_predictions} shows a strong tendency towards over-extraction, but also that some KPIs have been missed by the annotator. 
These unannotated but valid predictions demonstrate that ECB-A serves as a high-quality but inherently partial gold benchmark.

\subsection{Error Analysis}
Table \ref{tab:kpi_errors} shows the extraction most commonly evaluated as wrong. 
The top 2 ("azure ai VAL" and "1 billion") are meaningless labels. 
More interestingly 100\% of the extractions of cash flow are evaluated as wrong.
It seems from there the error rate drops quickly.
\begin{table}[htbp]
\centering
\footnotesize 
\setlength{\tabcolsep}{3pt} 
\begin{tabular}{p{4.0cm}rrr}
\toprule
\textbf{KPI (Centroid Label)} & \textbf{Total} & \textbf{Wrong} & \textbf{Err (\%)} \\
\midrule
azure ai VAL azure ai customers                     &  6 & 6 & 100.0 \\
1 billion                                           &  5 & 5 & 100.0 \\
cash flow                                           &  4 & 4 & 100.0 \\
rotcce                                              &  3 & 3 & 100.0 \\
electric utilities infrastructure up VAL            &  4 & 3 &  75.0 \\
international segment international segment revenue &  4 & 3 &  75.0 \\
nii                                                 &  8 & 4 &  50.0 \\
google service google service revenues              &  9 & 4 &  44.4 \\
organic sales growth                                & 15 & 4 &  26.7 \\
gross margin                                        & 17 & 4 &  23.5 \\
\bottomrule
\end{tabular}
\caption{Top 10 KPIs flagged by annotators as wrongly extracted, sorted by error rate.}
\label{tab:kpi_errors}
\end{table}
\paragraph{Differences between Calls and Filings} 
Why is there such a discrepancy between results on Earnings calls and SEC filings?
In this section, we highlight concrete differences between these two sources of information, with a thorough analysis of the difficulties for the KPI-extraction models. 
We use \textcolor{Green}{Green} for \colorbox{green!20}{KPI label} and \textcolor{blue}{Blue} for \colorbox{blue!20}{Value}.
\begin{tcolorbox}[colback=earnings_call,
  left=2pt,                 
  right=2pt,
  top=2pt,
  bottom=2pt]
As you know, \colorbox{green!20}{free cash flow} has been our primary financial metric through this recovery, and based on our
performance \colorbox{green!20}{year-to-date}, we still plan to be in the \colorbox{blue!20}{guidance range} for the year as well as the \colorbox{blue!20}{\$10 billion} target by \colorbox{green!20}{2025} and \colorbox{green!20}{2026}.
    \cite{Boeing_sec_2024}
\end{tcolorbox}
\noindent This statement by Boeing's CEO Dave Calhoun exemplifies the linguistic complexity of earnings calls, presenting challenges for natural language processing systems. 
The \textit{free cash flow} metric serves as the anchor for the ``\$10 billion target'', yet this relationship is obscured because this figure points to future performance (2025-2026) rather than the current period.
Meanwhile, current performance is described only vaguely through indirect reference to an unspecified \textit{guidance range}.
This linguistic structure creates an ambiguity pattern typical of earnings calls, in which optimistic numerical projections receive prominence while potentially damaging current KPIs remain underspecified. 
With respect to performance, \textit{Gemma-3} is able to do this perfectly, relating them with the label ``free cash flow 2025 2026'', the same is \textit{Gemini-3-pro} with the label ``plan target free cash flow 2025 and 2026''.
\textit{Llama-3.3} extracts both \$10 billion and free cash flow as entities; however, it does not relate them to each other in a group with a label.
Finally, \textit{Qwen} relates ``2025'', ``2026'', ``year-to-date'', altogether, and ends up using the label "still plan free cash flow guidance range year-to-date" with the \$10 billion as value.
This non-agreement between any of the models of course also means that our final system predicts nothing for cash flow guidance.
The omission of `free cash flow' from SEC filings---despite its status as a primary metric---exemplifies the unique value and challenge of the under-investigated earnings call domain. 
Our system successfully identified this as a longitudinal KPI across eight periods; notably, all four models correctly extracted the current `\$310 million' value with only minor label variations. 
However, longitudinal consistency in other quarters was occasionally disrupted by temporal ambiguity arising especially from the Q\&A session.

\paragraph{The High Variability in Call Culture}
The next quote highlights how much call culture varies across companies. 
\textsc{JP Morgan Chase} almost exclusively uses traditional performance measures in their calls.
They follow a strict format, where they read aloud their earnings material, resulting in more extractions by the SEC-based models.
Given the variability in call cultures, future studies should focus on scaling this benchmark to more companies.
\begin{tcolorbox}[colback=earnings_call,
  left=2pt,                 
  right=2pt,
  top=2pt,
  bottom=2pt]
     Starting on page 1, the Firm reported \colorbox{green!20}{net income} of \$\colorbox{blue!20}{12.9 billion}, \colorbox{green!20}{EPS} of \$ \colorbox{blue!20}{4.37} on revenue of \$\colorbox{blue!20}{43.3 billion} with an \colorbox{green!20}{ROTCE} of \colorbox{blue!20}{19\%}
\citep{JPMorgan2024EarningsCall}
\end{tcolorbox}
\paragraph{Q\&A Session}
Earnings calls usually end with a more informal Q\&A session. 
We examine the same example call featured in Figure \ref{fig:frontpage}, which demonstrates the huge value of mastering this domain. 
In the following quotes, we highlight modeling requirements unique to the Q\&A.
\begin{enumerate}
    \setlength{\itemsep}{-5pt} 
    \item Lyft does not make clear that it is a correction in their presentation, showing the need to keep track of \emph{inconsistencies}. 
    \item Understanding that multiple people are speaking, and how they are related to the call.
    \item Detection of negation, as Nikhil's mention of 500 basis points reflects what he believes the figure is \emph{not}.
    \item Detection of levels of abstraction, as the same metric can be referenced with different levels of specificity - e.g. `margin expansion' vs. `EBITDA margin expansion'.
\end{enumerate}
\begin{tcolorbox}[colback=Q_A,  left=2pt,
  right=2pt,
  top=2pt,
  bottom=2pt]
\textbf{Question from:} Nikhil Devnani (Analyst) \\
    ``Can we just please clarify the \colorbox{green!20}{EBITDA margin expansion}? I think the slide says \colorbox{blue!20}{500 basis points}. ... 
    But Erin, you mentioned \colorbox{blue!20}{50}. So, I think it is \colorbox{blue!20}{50}, but if you could just clarify that again, please?''
    \cite{MotleyFool2024LyftEarnings}
\end{tcolorbox}
\begin{tcolorbox}[colback=Q_A,  left=2pt,                 % text margin
  right=2pt,
  top=2pt,
  bottom=2pt]
\textbf{Answer from:} Erin Brewer -- CFO \\
``Thanks, Nikhil. This is Erin. And this is actually a correction from the press release. You're correct in my prepared remarks, I referenced \colorbox{blue!20}{50 basis points} of \colorbox{green!20}{
margin expansion''}
\cite{MotleyFool2024LyftEarnings}
\end{tcolorbox}

\section{Discussion}
Our empirical results and qualitative analysis reveal the high variability between earnings calls.
While market reaction to earnings reports is immediate and automated (Figure \ref{fig:frontpage}), the conversational nature and complexity of earnings calls have so far prevented similar high-speed, autonomous absorption.
It is clear that earnings calls, down to the level of specific KPIs, influence the stock price. 
BERT-based financial KPI extraction models function well for SEC filings, as they are standardized by strict auditing and close to devoid of cultural variation. 
However, they fail to generalize to the subjective and promotional language used in earnings calls, posing challenges for current NLP methods.
This complexity is compounded by variability in company-specific styles and cultural factors inherent to each organization or, indeed, each speaker.
Our analysis and system provide insights into the challenges and opportunities in earnings calls, and while our human evaluation reveals an error rate necessitating human oversight, it is a step towards more efficient and faster processing of new information.
Our exploration of KPIs present in earnings calls lays the groundwork for fine-tuning large language models for extractions.

\section{Conclusion}
This work characterizes the unique challenges and opportunities in automated KPI extraction from earnings calls.
We introduce three novel benchmarks: the SEC filings benchmark (\textbf{SECB}), the earnings calls benchmark (\textbf{ECB}) with a smaller annotated subsample (\textbf{ECB-A}). 
Our empirical evaluation demonstrates that, while current KPI extraction methods show generalization capabilities across SEC filings datasets, they do not generalize to the more unstructured nature of earnings calls. 
Our qualitative analysis reveals why earnings calls present unique challenges, complicating KPI extraction due to subjective phrasing, company-specific terminology, and varying levels of formality that contrast sharply with structured SEC filings. 
Finally, validated by human evaluation, our work provides a robust baseline for the emerging task of automated KPI extraction from this valuable data source, with experiments and analysis laying the groundwork for future advances in real-time financial decision-making.

\section*{Limitations}
\paragraph{Data Scale and Annotation}
Due to the scarcity of experts in this domain and the compensation such experts usually demand, we were only able to recruit one annotator for the ECB-A dataset. 
We try to mitigate this by having the expert go over the annotations twice.
Even though we attempt to get as diverse a sample as possible by randomly sampling 10 distinct companies in various industries, as mentioned in the paper, cultural differences between companies mean that we do not necessarily know how well these results generalize, especially outside of major US companies.

\paragraph{Methods and Evaluation}
The selection of a cutoff of 0.85 for semantic similarity was empirically derived, even though we provide a sensitivity analysis in the appendix \ref{Sensitivity_hyperparam} showing that our results are stable to some degree of change in parameters. 
Ideally, future work should aim to have this dynamically tuned by the actual model using clustering, e.g., K-means or, likely better-suited for this task, DBSCAN.
There is a potential issue with data leakage, which could be and most likely is part of the training data for some of these LLMs.
\paragraph{Evaluation}
Some of the evaluation relies on an LLM as a judge, where it is important to note that an LLM as a judge is not always reliable; we try to mitigate this by using other automatic metrics as well as human evaluators.
Because it is significantly easier to verify a correct result than annotate a ground truth, we utilize 3 human judges, who, however, must be noted as not experts in the field, though with a basic understanding.

\section*{Ethics statement}
There are risks with automated system especially in a financial context. There is a system risk that wrong extraction could result in the wrong financial decision temporaily pricing a stock at the wrong price. 
Which could lead to financial losses and gains for other actors in the market as well as the system user.
There is a risk that systems like these put institutional investors even further in front of retail investors with less sophisticated setups for investment; however, faster, accurate pricing of securities also has the advantage of less volatility in the markets, as well as fairer prices.
Our work is based on readily available data and adheres to the ACL Code of Ethics.

\section*{Acknowledgments}
We would like to thank the AAU-NLP group for helpful discussions and feedback on an earlier version of this article. 
We would like to give a special acknowledgement to Ernests Lavrinovics, for helping with evaluation of our final system. 
We want to also thank Alipes ApS for their support in facilitating and funding this research and as well as useful discussions with their Quant team.
Rasmus Aavang is supported by the Industrial Ph.D. programme from Innovation Fund Denmark (grant code 4297-00016B).
MZ and JB, were supported by the research grant (VIL57392) from VILLUM FONDEN. 
MZ also received funding from the Danish Government to Danish Foundation Models (4378-00001B).

%\section*{Acknowledgments}
% TODO remember to highlight Ernest for his help in evaluation.
% Innovation Fund Denmark
% Reviewer comments
% Anyone in AAU-NLP in particular

\bibliography{custom}

\appendix
\clearpage
\section{Detailed Experimental Setup}
\label{app:detailed_setup}
\subsection{Model setup}
Following the explanation in \citep{loukas-etal-2022-finer}, we download their SEC-BERT-BASE model from Hugging face \citep{huggingface2024}, we use Hugging face to download their dataset and then train the model on the FiNER-139 train set, with the validation set as evaluation on a GTX 1080 TI for 10 epochs with early stopping patience of 2 with a batch-size of 32 (not specified in the \citep{loukas-etal-2022-finer} paper), and set it to truncation at max length of 512 tokens (also not specified) to match the limit for the input size of the SEC-BERT-BASE model.
The early stopping performance improvement was also based on the validation set of FiNER-139, the best model was achieved after all 10 epochs had run.

\subsection{Parser}
\label{app:parser}
The parser drops all tables, then tries to grab text inside ['p', 'div', 'span', 'section'], then, having done that, we deduplicate based on which extraction leads to the longest substring.
After this, we do postprocessing, cleaning out malformed snippets in the same way as \cite{aavang2025hifi}, where we check if it starts with a "." and clean any potential leading whitespace, lastly dropping snippets not starting with a capital letter.
Furthermore, we drop all snippets more than 3 std. deviations longer than the mean, meaning snippets longer than 4513 characters.

\subsection{Standardizing the Label Space between SEC-based models.}
One of the core issues we had to figure out to compare the different SEC filings datasets across different datasets is how they are majorly finetuned to a certain dataset. 
Therefore, the labelset varies significantly in the number of unique labels present.
\begin{table}[t]
    \centering
    \begin{tabular}{lr} 
        \toprule
        \textbf{Dataset} & \textbf{Unique Labels} \\
        \midrule
        FiNER-139     & 140 \\
        HiFi-KPI      & 198K \\
        \textbf{SECB} & 1,615 \\
        \bottomrule
    \end{tabular}
    \caption{Unique labels in each dataset.}
    \label{tab:unique_labels}
\end{table}
We therefore have to convert between these sets we do this in the following way.
First since the formatting of the labels are slightly different we cut out the "us-gaap" part of the SL1000 model predictions if predicting on the FiNER-139 dataset.
Then we check if a label with the same name is part of the set and if not we give the pseudo label "UNK" that we can then use to track if it is at least right in the broader context of something being a fincial key figure or not.
When it comes to the conversion the other way around where SEC-BERT-BASE does not have the label that is actually present if it predicts something for a label that then has a label not part of the SEC-BERT-BASE label space we then use the placeholder "UNK" as well.
An interesting thing for this is that it means two things are true. 
None of the models can ever correctly predict the regex\_label / regex\_percentage labels. 
The special-OOS label that the SL1000 based models all can also never be completely correct.

\begin{table}[t]
    \centering
    \begin{tabular}{lr} 
        \toprule
        \textbf{Dataset} & \textbf{Count} \\
        \midrule
        SECB       & 21,258 \\
        HiFi-KPI   & 159,481 \\
        FiNER-139  & 40,569 \\
        \bottomrule
    \end{tabular}
    \caption{BERT-SL1000 use of the special-OOS.}
    \label{tab:placeholder}
\end{table}

\subsection{ECB \& ECB-A}
We segment the calls by looking for the occurrence of newlines in connection with a speaker's name. If a newline occurs before a speaker's name, we assume it to be a new speaker.
Furthere we drop all transcriptions of what the operator says in the call as they will not say anything interesting or containing KPIs anyway.
Finally, to manage context size for the bert based models, we further split and reconstruct the chunk with spaCy if any chunk contains more than 10 sentences.

\section{Matching logic for automatic evaluation of ECB-A}
\begin{comment}

\begin{table}[]
    \centering
    \begin{tabular}{c|c}
         &  \\
         & 
    \end{tabular}
    \caption{Normalization used in matches}
    \label{tab:placeholder}
\end{table}
\end{comment}
To calculate automatic metrics, we take any direct supersets of both the model predictions and the ground truths for a certain chunk, and discard if there is anyway is any set tagged that is just the more elaborate label with less information.
We then take all the candidate extractions and ground truth values then we compare these against each other. 
The best scoring extraction then consume the ground truth match.
Meaning that there is a one-to-one mapping, this is such that a model can not artificially enhance its score by predicting the same correct label many times.
The value match is successful if ground truth and values either have the same value or, if they match the value, are multiplied by a different multiple of 1000s and have a cross-encoder score over 0.75. We count it as a value match as well.
If the model has extracted something with the is\_range parameter as true, then we allow it to potentially consume multiple ground truths, such that if, e.g. "4-5 unit a month" it can match both 4 and 5.
Finally, for non-numeric values (e.g., 'Record'), we consider these a match if the Gestalt pattern matching similarity ratio between the extraction and ground truth strings are greater than 0.8.
\section{Sensitivity in Threshold for the Semantic Clustering}
\label{Sensitivity_hyperparam}

\begin{table}[h]
\centering
\footnotesize	
% \resizebox matches the table width to the column width
\resizebox{\columnwidth}{!}{%
    \begin{tabular}{lrrr}
    \toprule
    \textbf{Model} & \textbf{Share} & \textbf{Centroid} & \textbf{Overlap} \\
     & \textbf{(\%)} & \textbf{(\%)} & \textbf{(\%)} \\
    \midrule
    Llama-3.3-70B & $77.50 \pm 0.66$ & $62.51 \pm 9.56$ & $31.03 \pm 1.21$ \\
    Qwen3-30B-A3B & $78.70 \pm 0.03$ & $61.61 \pm 8.54$ & $25.49 \pm 1.56$ \\
    Gemma-3-27B   & $77.58 \pm 0.33$ & $58.93 \pm 8.59$ & $30.10 \pm 1.11$ \\
    \bottomrule
    \end{tabular}%
}
\caption{\textbf{Variance in model contributions.} Mean and standard deviation across the 4 different parameters setting. 
The average total number of extractions (dataset size) was $1354.75 \pm 193.34$.}
\label{tab:model_variance}
\end{table}

Table \ref{tab:model_variance} show the robustness of our threshold for the clustering cutoff by trying [0.75, 0.80, 0.85, 0.9] as well. 
We see that it doesn't have a big impact on the final clusters; however, there are, of course less clusters the higher you set the threshold.

\section{ECB-A Annotation Setup}
\begin{figure*}[t]
    \centering
    \includegraphics[trim=0.2cm 1cm 12cm 1cm, clip, width=1\linewidth]{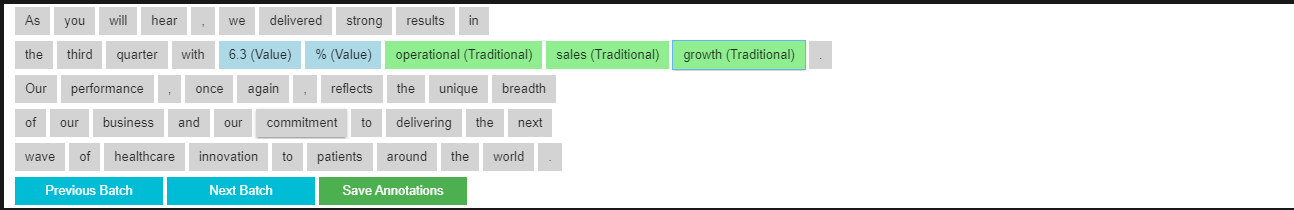}
    \caption{\textbf{Tagging Interface During Annotation} Example : 2024 Q3 JNJ earnings transcript}
    \label{fig:interface}
\end{figure*}
\begin{figure*}[h]
    \includegraphics[width=\textwidth]{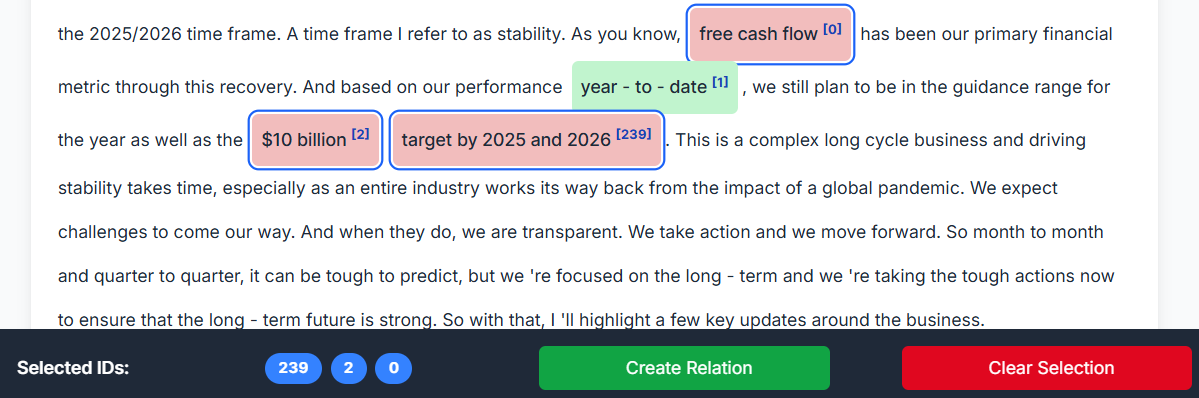}
    \centering
    \caption{\textbf{Relation Extraction Annotation Interface.}}
    \label{fig:relationExtraction}
\end{figure*}

\label{ManualAnnoSetup}
The annotation resembles the idea behind \citep{hillebrand2022kpibertjointnamedentity}
The expert annotator was put in a setting where they had to review at least the whole earnings transcript in one sitting to try to balance for fatigue in the mundane task across companies. 
They were given the possibility of using 3 options "Traditional", "Non-traditional", and "Value". 
\begin{table}[h]
    \centering
    \begin{tabular}{|c|c|}
    \hline
        \cellcolor{green!20} One click  & \cellcolor{green!20} Traditional\\ \hline
        \cellcolor{Orange!20} Two click & \cellcolor{Orange!20} Non-Traditional \\ \hline
        \cellcolor{blue!20} Third click & \cellcolor{blue!20} Value \\ \hline
        \cellcolor{gray!20} Fourth click & \cellcolor{gray!20} Reset \\ \hline
    \end{tabular}
    \caption{Interface for manual annotator}
    \label{tab:my_label}
\end{table}
They could then click on the interface in figure \ref{fig:interface}.
One click meant Traditional, another click "non-traditional" yet another "value" and a fourth click resets the annotation of the token.
They were free to go back and forth by themselves during the tagging process and correct their annotations.
%Where in the calls the annotations fall can be seen in Figure \ref{fig:Distribution}
\\
Then, at a later date, the annotations were confirmed by the same annotator in another interface, where one would be able to model relations between entities. Here, the annotator was instructed to annotate related entities, enabling later relation extraction.
This was done by the interface in Figure \ref{fig:relationExtraction} where you click on each entity and then create a relation.
\subsection{Traditional vs Non-Traditional Performance metrics}
\begin{table}[h!]
\centering
\scriptsize 
\setlength{\tabcolsep}{3pt}
\begin{tabular}{p{0.24\textwidth} | p{0.24\textwidth}}
\toprule
\textbf{Traditional Performance Measures} & \textbf{Non-traditional Performance Measures} \\ \midrule
Based on outdated traditional accounting system         & Based on company strategy                     \\
Mainly financial measures                               & Mainly non-financial measures                \\
Intended for middle and high managers                  & Intended for all employees                   \\
Lagging metrics (weekly or monthly)                    & On-time metrics (hourly, or daily)           \\
Difficult, confusing, and misleading                   & Simple, accurate, and easy to use            \\
Lead to employee frustration                           & Lead to employee satisfaction                \\
Neglected at the shopfloor                             & Frequently used at the shopfloor             \\
Have a fixed format                                    & Have no fixed format (depends on needs)      \\
Do not vary between locations                         & Vary between locations                       \\
Do not change over time                               & Change over time as the need changes         \\
Intended mainly for monitoring performance            & Intended to improve performance              \\
Not applicable for JIT, TQM, CIM, FMS, RPR, OPT, etc. & Applicable                                   \\
Hinders continuous improvement                        & Helps in achieving continuous improvement    \\ \bottomrule
\end{tabular}
\caption{Comparison of Traditional and Non-traditional Performance Measures from \cite{ghalayini1996changing}}
\label{tab:Tradional}
\end{table}
Table \ref{tab:top-kpis} shows the most common labels annotated as either tradtional or non-traditional in ECB-A.
\section{Evaluation}
\label{Evaluation}
\begin{figure}[t]
    \centering
    \includegraphics[width=\columnwidth]{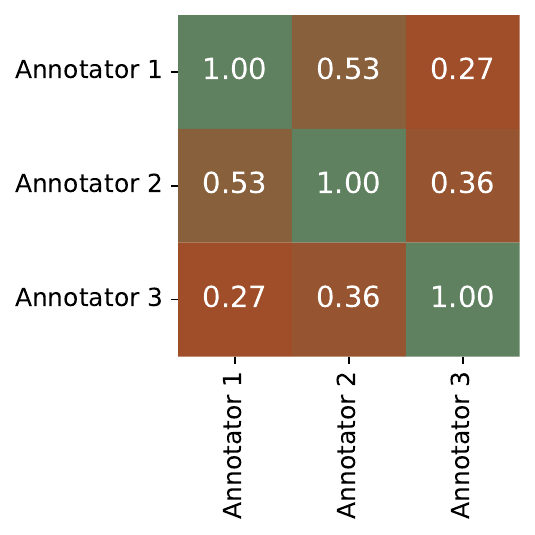}
    \caption{\textbf{Inter-Annotator Agreement (Cohen's Kappa).} 
    Annotators 1 and 2 exhibit strong alignment, whereas Annotator 3 demonstrates notably lower agreement with the other evaluators.}
    \label{fig:cohens_kappa}
\end{figure}
The annotation setup for the evaluation of the final system consists of 3 annotators, each were tasked with annotating 200 extractions each, with 100 extractions overlapping between the annotators.
The setup was a command-line tool built in Python, that presented the annotators with the extracted KPI Label and value; they were then to evaluate if this extraction was correct or not from the corresponding chunk the system had extracted the KPI from. 
They could then either input yes or no, and if they selected no, they could present a short reasoning for why they did not think the extraction was correct.
Full annotator guidelines for evaluation in \ref{Lst:evaluation} and the accompanying table specifying fiscal years in Table \ref{tab:company_fiscal_periods}.
The Cohen's kappa between annotators can be seen in Figure \ref{fig:cohens_kappa}

\begin{table*}[t]
    \centering
    \begin{tabular}{lllllll}
        \toprule
        \textbf{Ticker} & \textbf{Company} & \textbf{FY End} & \textbf{Q1} & \textbf{Q2} & \textbf{Q3} & \textbf{Q4} \\
        \midrule
        AAPL & Apple Inc. & Late Sep & Oct--Dec & Jan--Mar & Apr--Jun & Jul--Sep \\
        HD & Home Depot & Late Jan & Feb--Apr & May--Jul & Aug--Oct & Nov--Jan \\
        MSFT & Microsoft Corp. & Jun 30 & Jul--Sep & Oct--Dec & Jan--Mar & Apr--Jun \\
        PG & Procter \& Gamble & Jun 30 & Jul--Sep & Oct--Dec & Jan--Mar & Apr--Jun \\
        AMZN & Amazon.com Inc. & Dec 31 & Jan--Mar & Apr--Jun & Jul--Sep & Oct--Dec \\
        BA & Boeing Co. & Dec 31 & Jan--Mar & Apr--Jun & Jul--Sep & Oct--Dec \\
        BAC & Bank of America & Dec 31 & Jan--Mar & Apr--Jun & Jul--Sep & Oct--Dec \\
        CAT & Caterpillar Inc. & Dec 31 & Jan--Mar & Apr--Jun & Jul--Sep & Oct--Dec \\
        CVX & Chevron Corp. & Dec 31 & Jan--Mar & Apr--Jun & Jul--Sep & Oct--Dec \\
        DOW & Dow Inc. & Dec 31 & Jan--Mar & Apr--Jun & Jul--Sep & Oct--Dec \\
        GOOGL & Alphabet Inc. & Dec 31 & Jan--Mar & Apr--Jun & Jul--Sep & Oct--Dec \\
        JNJ & Johnson \& Johnson & Dec 31 & Jan--Mar & Apr--Jun & Jul--Sep & Oct--Dec \\
        JPM & JPMorgan Chase & Dec 31 & Jan--Mar & Apr--Jun & Jul--Sep & Oct--Dec \\
        KO & Coca-Cola Co. & Dec 31 & Jan--Mar & Apr--Jun & Jul--Sep & Oct--Dec \\
        NEE & NextEra Energy & Dec 31 & Jan--Mar & Apr--Jun & Jul--Sep & Oct--Dec \\
        PFE & Pfizer Inc. & Dec 31 & Jan--Mar & Apr--Jun & Jul--Sep & Oct--Dec \\
        PLD & Prologis Inc. & Dec 31 & Jan--Mar & Apr--Jun & Jul--Sep & Oct--Dec \\
        XOM & Exxon Mobil & Dec 31 & Jan--Mar & Apr--Jun & Jul--Sep & Oct--Dec \\
        \bottomrule
    \end{tabular}
    \caption{Fiscal year end dates and quarterly periods.}
    \label{tab:company_fiscal_periods}
\end{table*}

\onecolumn
\subsection{Evaluator Guidelines}
\begin{lstlisting}[
    basicstyle=\scriptsize\ttfamily,
    breaklines=true,
    breakatwhitespace=true,
    frame=single,
    caption={The annotator guidelines we followed in annotation of the transcripts.},
    captionpos=b,
    label={Lst:evaluation}
]
1. Task Overview
The goal of this task is to accurately identify and verify financial data points within company transcripts. 
2. General Labeling Principles
Strict Extraction: All components of a label must be present in the actual text. Do not paraphrase.
Range Handling: If a value falls in the middle of a provided range in the text, it should be marked as correct.
Abbreviations: Be aware that standard financial abbreviations are frequently used. 
e.g.:
Opex: Operating Expenses
Capex: Capital Expenditure
3. Financial Terminology & Unit Conversions
Basis Points (bps). 
Conversion Table: 
1 Basis Point | 1 bp | 0.01% | 0.0001 | 1/10,000 
100 Basis Points | 100 bps | 1.00% | 0.01 | 1/100
4. Metadata & Period Verification
You will be provided with metadata at the top of the task to identify the source.
Example Header: Company: AAPL | Year: 2024 | Quarter: 4
Model/system Date: Date: 2024-Q4 (This is a system-generated field).
Fiscal Year vs. Calendar Year: please verify that the transcript references the correct period in the data.
Do not assume that Q1 is always January to March
Many companies have Fiscal Years (FY) that do not align with the Calendar Year.
Example: A company's Q1 might run from July to September.
Action: you can refer to the provided Fiscal Calendar Table to confirm the specific reporting period for the company in question.
5. Error Handling & Rejections
If you identify a label as incorrect, you must provide a justification.
Requirement: When rejecting a label, write a clear, concise reason explaining why it is invalid (e.g., "Wrong time period," "Value not in text," "Hallucinated number").
\end{lstlisting}

\subsubsection{Full SECB Example}
\begin{lstlisting}
    

{
   "form_type": "10-K",
   "accession_number": "0000012927-23-000007",
   "filing_date": "20230127142633",
   "quarter_ending": "20221231",
   "company_name": "BOEING CO",
   "text": "The Company's deferred income tax assets of $12,301 can be 
   used in future years to offset taxable income and reduce income 
   taxes payable. The Company's deferred income tax liabilities of 
   $9,306 will partially offset deferred income tax assets and result 
   in higher taxable income in future years and increase income taxes
   payable. Tax law determines whether future reversals of temporary 
   differences will result in taxable and deductible amounts that 
   offset each other in future years. The particular years in which 
   temporary differences result in taxable or deductible amounts 
   generally are determined by the timing of the recovery of the 
   related asset or settlement of the related liability. The deferred 
   income tax assets and liabilities relate primarily to U.S. federal
   and state tax jurisdictions. From a U.S. federal tax perspective
   , the Company generated a tax NOL in 2020 that was carried back to 
   prior years when the tax rate was 35% due to the CARES Act benefit 
   as described above. The Company generated tax NOL in 2021 and 
   interest carryovers in 2021 and 2022 that can be carried forward
   indefinitely and federal research and development credits that can
   be carried forward 20 years.",
"entities": [
 [
   45,
   51,
   "us-gaap:DeferredTaxAssetsGross",
   "instant",
   "2022-12-31",
   "2022-12-31",
   "iso4217:USD",
   12301000000.0
 ],
 [
   188,
   193,
   "us-gaap:DeferredIncomeTaxLiabilities",
   "instant",
   "2022-12-31",
   "2022-12-31",
   "iso4217:USD",
   9306000000.0
 ],
 [
   938,
   940,
   "us-gaap:EffectiveIncomeTaxRateReconciliation
   AtFederalStatutoryIncomeTaxRate",
   "duration",
   "2020-03-27",
   "2020-03-27",
   "xbrli:pure",
   0.35
 ]
]
}
\end{lstlisting}

\subsection{Annotator Guidelines}
\begin{lstlisting}[
    basicstyle=\scriptsize\ttfamily,
    breaklines=true,
    breakatwhitespace=true,
    frame=single,
    caption={The annotator guidelines we followed in annotation of the transcripts.},
    captionpos=b,
    literate={“}{{``}}1 {”}{{''}}1,
    label={Lst:Annotator}
]
Basic guidelines:
We want to find KPI_FACTS in the text. To find these, you are going to do two things.
Tag entities in the text
Relate entities from (1) to each other.

The entities are broadly the following:
KPI 
Scope
Value 
Temporal Context or Modifier
Modifiers 

Note that there are likely gonna be a lot of cases where not all categories are present
Follow these steps for every sentence:
Read for Understanding: First, read the entire text snippet to understand its full meaning.
Tag All Entities: Identify and label the five entity types in the sentence. Be precise with your highlighting (span selection).
Draw All Relations: Connect the entities according to the rules in Section 5. A KPI is the central hub for all relations.
Numerical Values: 
Annotate all the number values that appear; these can be in the form of dollar figures, percentages, records, or the number of products produced.
Please annotate the $ sign as well as any order of magnitude specifier 
E.g. ,"$5 million", "five million dollars"
Soft (Vague or qualitative) Values:
“For the year, we saw top-line growth from rate cases and riders across our jurisdictions.”

Please also annotate more soft forms of KPIs, such as “thousands of units”,  “record number”, “strong growth”, “stable rate” etc.

“strong growth in YouTube subscriptions”
“will reflect the increases in depreciation and expenses”
Increase -> value 
depreciation and expenses ->  KPI.

KPI descriptions:
Having tagged the value, you should tag the related description.
The name of the metric. This is the core concept you will be annotating.
Rule: Tag the complete noun phrase that defines the metric. This must include essential modifiers that change the metric's definition.
Include: Adjectives like net, gross, adjusted, quarterly, annual, monthly, recurring.
Exclude: Determiners (a, the, our), verbs (reached, was), and descriptive but non-essential words (strong, record) if a KPI_VALUE is present

This is for example:
“quarterly net revenue” -> whole company description
“Google Services operating margins” -> that subdivision
“operating margins” -> whole company 

Example sentence:
“[retention] is dropping down to [70%]”
“Retention “-> KPI
70%->  value
Modifiers:
Do include Essential modifiers, e.g., “net”, “gross”, “adjusted”, “quarterly”, ”annual”

Example “Our [quarterly net revenue], driven by strong performance in the [cloud division], reached a [record] [$10B]” 

See clarification for the “cloud division” in the following.
Scope - Subcomponent or Product:
If some KPI has to do with some specific subcomponent of the company, please be sure to tag that accordingly as well. Here, it is common for companies to go through a whole subdivision at once, meaning you often have to relate the subdivision to many KPIs.

Business division A formal part of the company's structure e.g., “Cloud Division”, “Services Arm”, “Global Operations”
A Product or Service: A specific offering from the company (e.g., “iPhone”, “Windows 11”, “Model 3”).
A Geographical Market: A region where the company operates (e.g., “North American”, “Asia”).

Temporal context / Modifier:
Please also connect with the temporal description, this can be “2025”, “2026”, “next year”, “in the current quarter”, “year to date” and many more

For temporal context, please make sure to also annotate if something is “expected”, “target”, “projected”, “expected”
Relation:
You should use the annotation tool to relate the KPI description to its values; you have the “value_of” relation
Such that each group describes one direct connection between value and KPI and potential modifiers.
Examples:
Our [quarterly net revenue], driven by strong performance in the [cloud division], reached a record [$10B]

[quarterly net revenue]<->[cloud division]<-> [$10B] => (value_of)

“Our [guidance] for [Q1 2026] is [projected] [quarterly net revenue] from the [Cloud Division] of [$12B]."

Tagged Entities
[guidance] - MODALITY
[Q1 2026] - TEMPORAL_CONTEXT
[projected] - MODALITY
[quarterly net revenue] - KPI
[Cloud Division] - SCOPE
[$12B] - KPI_VALUE

Relation These would all be related in the same group:
[guidance]  [Q1 2026] [projected] [quarterly net revenue] [Cloud Division]  [$12B]

Multiple entities:

"[Cloud Division] [revenue] was [$10B] in [2024], while [Services Arm] [profit] is [expected] to be [$3B]."

In this case, there are two distinct facts about two different KPIs. You would create two separate KPI relations groups
KPI_Fact 1:
KPI: [revenue]
KPI_VALUE: [$10B]
SCOPE: [Cloud Division]
TEMPORAL_CONTEXT: [2024]
KPI_Fact 2:
KPI: [profit]
KPI_VALUE: [$3B]
SCOPE: [Services Arm]
MODALITY: [expected]
\end{lstlisting}
\twocolumn

\section{Cost \& Runtime}
Table \ref{tab:time_cost_metrics} shows both the runtime and the cost of running these large LLM-based models of running the different LLMs on the Apple Q1 2023 earnings call transcript, which ends up being 54 chunks.
Gemma-3-27B is actually free (for us) because we use Google's free endpoint.
The runtime also shows that from a financial perspective there is still gains to be made just from being faster.
It is also clear that the slower the model the better performance.

\begin{table*}[ht]
\centering
\begin{tabular}{lrrrrrr}
\toprule
& \multicolumn{3}{c}{\textbf{Time Metrics (s)}} & \multicolumn{3}{c}{\textbf{Cost Metrics (\$)}} \\
\cmidrule(lr){2-4} \cmidrule(lr){5-7}
\textbf{Model} & \textbf{Mean} & \textbf{Std Dev} & \textbf{Total} & \textbf{Mean} & \textbf{Std Dev} & \textbf{Total} \\
\midrule
Gemma-27b-it           & 15.06 & 22.32 &  813.04 & 0.000000 & 0.000000 & 0.000000 \\
Qwen3-30b-a3b          &  5.94 & 18.18 &  320.67 & 0.000418 & 0.000584 & 0.022553 \\
Llama-3.3-70b-instruct &  5.72 & 17.07 &  308.65 & 0.002075 & 0.001187 & 0.112056 \\
Gemini-3-pro preview   & 48.22 & 37.29 & 2603.93 & 0.049013 & 0.036039 & 2.646720 \\
\bottomrule
\end{tabular}
\caption{Comparison of execution time and API costs across evaluated models.}
\label{tab:time_cost_metrics}
\end{table*}
\begin{table*}[t]
\centering
\begin{tabular}{lr}
\toprule
\textbf{KPI Label} & \textbf{Unique Companies} \\
\midrule
revenue & 5 \\
free cash flow & 4 \\
net income & 4 \\
capex & 3 \\
cash flow & 2 \\
\bottomrule
\end{tabular}
\caption{Top 5 most common KPIs consistently reported across unique companies.}
\label{tab:kpi_coverage}
\end{table*}
\begin{table*}[ht]
\centering
\begin{tabular}{clclc}
\toprule
\textbf{Rank} & \textbf{Traditional KPI} & \textbf{Occurrences} & \textbf{Non-Traditional KPI} & \textbf{Occurrences} \\
\midrule
1 & Revenue & 10 & Active Devices & 2 \\
2 & EPS & 6 & Apple Pay Available & 2 \\
3 & Operational Sales Growth & 6 & 737S Production Deliveries & 2 \\
4 & Operating Margin & 5 & FDA Approval & 2 \\
5 & Revenues & 5 & Freeform, A Brand-New App & 1 \\
\bottomrule
\end{tabular}
\caption{Top 5 most common Traditional and Non-Traditional KPIs.}
\label{tab:top-kpis}
\end{table*}

\section{Prompt Setup Details}

\label{app:prompt}
We use Openrouter\footnote{See \url{https://openrouter.ai/}.} to run Llama-70B, Gemini-3 pro and Qwen-3 model, we use Google Cloud\footnote {See \url{https://cloud.google.com/}} to run Gemma model and we use the OpenAI Python API library.
For the models that support an extraction schema, we utilize the extraction schema in \ref{ExtractionSchema}, and for the models that do not, we only utilize the prompt in \ref{Lst:oneshot}

\onecolumn
\begin{lstlisting}[
    basicstyle=\scriptsize\ttfamily,
    breaklines=true,
    breakatwhitespace=true,
    frame=single,
    caption={The few-shot prompt used.},
    captionpos=b,
    literate={“}{{``}}1 {”}{{''}}1, 
    label={Lst:oneshot}
]


You are an expert financial entity extractor. Your sole task is to read `### TEXT TO ANALYZE ###` section and extract all entities according to the JSON schema
    
    ## Entity Extraction Instructions ##
    Entities could be:

    * `kpi_name` The name of the metric.
        * **Rule:** Tag the complete noun phrase, including essential modifiers like "net", "gross", "adjusted", "quarterly", or "annual".
        * **Rule:** EXCLUDE non-essential fluff ("strong," "record") and determiners ("a," "the," "our").
        * Examples: "quarterly net revenue", "Google Services operating margins", "retention rate".

    * `kpi_value` The *quantifiable* value of the KPI.
        * **Rule:** This must be a numerical value.
        * **Math Rule:** If a range is provided (e.g., "$10-20M"), calculate the arithmetic average for the `Value` field, but record the bounds in the range fields.
        * Examples: "$10B", "70%", "five million dollars", "thousands of units".

    * `qualitative_desc`  A *subjective* or *non-numerical* description of the KPI's performance or trend.
        * **Rule:** This must highlight a specific qualitative milestone.
        * Examples: "strong growth", "stable rate", "increase", "dropping down", "disappointing results", "record number".

    * `scope` The specific business unit, product, or market the KPI refers to.
        * Examples: "Cloud Division", "Services Arm", "iPhone", "North American", "Services", "Boeing Commercial", "Boeing Defense and Space"

    * `date` The temporal context for the KPI.
        * **Rule:** This should include any relevant time frames or specific dates. Including if something is a future projection or historical fact.
        * Examples: "2024," "Q1 2026," "next year," "in the current quarter," "year to date", "end of year", "expect", "project"*

    * `modality` The certainty or context of the fact (e.g., if it's a projection vs. a reported fact).
        * **Rule:** Forward of backwards looking (e.g., "guidance" and "projected").
        * Examples: "projected", "expected", "target", "guidance".

    Having identified the entities, you should structure them into groups relating relevant entities together.

    ### 2. Field Definitions ###
    * **Source:** The exact text span from the input from which the metric was derived.
    * **Entities:** A list of all relevant entities in the text.
    * **Source Value:** The original text value of the metric.
    * **Label:** Construct a concise label from the entities in the text this could be `scope`, `kpi_name`, `date` and `modality`. Use only entities present in the source text.
    * **Value:** The numerical value as a float. If a range, use the average.
    * **Value_NonNumeric:** If the value is a non numerical highlight of performance.
    * **Is_Range:** Boolean indicating if the value comes from a range.
    * **Top_of_range / Bottom_of_range:** The specific upper/lower bounds if Is_Range is True.

    **Label Construction Rule:**
    You must generate a standard `Label` for each group to serve as a unique ID.
    * **Source:** Use ONLY the text of the entities found in that specific group.
    * **Order:** Construct the string in this exact precedence:
        1.  `scope`
        2.  `modality`
        3.  `kpi_name`
        4.  `date`
    * **Formatting:** Separate parts with a single space.
    * **Example:** If you find Scope="Cloud", KPI="Revenue", Date="Q1", the Label is "Cloud Revenue Q1".

    Example 1:
    "Quarterly revenues crossed the $10 billion mark for the first time"
    Extracted as
    {
    "Entities": [
            {"text": "Quarterly", "category": "date"},
            {"text": "revenues", "category": "kpi_name"},
            {"text": "$10 billion", "category": "kpi_value"}
        ],
    "Groups" : [{
        "Source": "Quarterly revenues crossed the $10 billion mark for the first time",
        "Entities": [
                {"text": "Quarterly", "category": "date"},
                {"text": "revenues", "category": "kpi_name"},
                {"text": "$10 billion", "category": "kpi_value"}
            ],
        "Source Value": "$10 billion",
        "Is_Range": false,
        "Top_of_range": null,
        "Bottom_of_range": null,
        "Value": 10000000000.0,
        "Value_NonNumeric": null,
        "Label": "revenues Quarterly"
    }]
    }

    Example 2:
    "Boeing Defense and Space. BDS booked $6 billion in orders during the quarter. Revenue was $5.5 billion"
    Extracted as
    {
    "Entities": [
        {"text": "Boeing Defense and Space", "category": "scope"},
        {"text": "BDS", "category": "scope"},
        {"text": "$6 billion", "category": "kpi_value"},
        {"text": "orders", "category": "kpi_name"},
        {"text": "during the quarter", "category": "date"},
        {"text": "Revenue", "category": "kpi_name"},
        {"text": "$5.5 billion", "category": "kpi_value"}
    ],
    "Groups" : [{
        "Source": "Boeing Defense and Space. BDS booked $6 billion in orders during the quarter",
        "Entities": [
                        {"text": "Boeing Defense and Space", "category": "scope"},
                        {"text": "BDS", "category": "scope"},
                        {"text": "orders", "category": "kpi_name"},
                        {"text": "during the quarter", "category": "date"},
                        {"text": "$6 billion", "category": "kpi_value"}
                    ],
        "Source Value": "$6 billion",
        "Is_Range": false,
        "Top_of_range": null,
        "Bottom_of_range": null,
        "Value": 6000000000.0,
        "Value_NonNumeric": null,
        "Label": "Boeing Defense and Space BDS orders during the Quarter"}
    ,
    {
        "Source": "Boeing Defense and Space. BDS booked $6 billion in orders during the quarter. Revenue was $5.5 billion",
        "Entities": [
                        {"text": "Boeing Defense and Space", "category": "scope"},
                        {"text": "BDS", "category": "scope"},
                        {"text": "Revenue", "category": "kpi_name"},
                        {"text": "$5.5 billion", "category": "kpi_value"},
                        {"text": "during the quarter", "category": "date"}
                    ],
        "Source Value": "$5.5 billion",
        "Is_Range": false,
        "Top_of_range": null,
        "Bottom_of_range": null,
        "Value": 5500000000.0,
        "Value_NonNumeric": null,
        "Label": "Boeing Defense and Space BDS Revenue during the Quarter"
    }]
    }

    Example 3:
    "We expect net income to be in the range of $1.2 billion to $1.4 billion for the fiscal year 2026."
    Extracted as
    {
        "Entities": [
            {"text": "expect", "category": "modality"},
            {"text": "net income", "category": "kpi_name"},
            {"text": "$1.2 billion", "category": "kpi_value"},
            {"text": "$1.4 billion", "category": "kpi_value"},
          {"text": "fiscal year 2026", "category": "date"}
        ],
        "Groups": [{
            "Source": "We expect net income to be in the range of $1.2 billion to $1.4 billion for the fiscal year 2026.",
            "Entities": [
                {"text": "expect", "category": "modality"},
                {"text": "net income", "category": "kpi_name"},
                {"text": "fiscal year 2026", "category": "date"},
                {"text": "$1.2 billion", "category": "kpi_value"},
                {"text": "$1.4 billion", "category": "kpi_value"}
            ],
            "Label": "expect net income fiscal year 2026",
            "Source Value": "$1.2 billion to $1.4 billion",
            "Value": 1300000000.0,
            "Value_NonNumeric": null,
            "Is_Range": true,
            "Top_of_range": 1400000000.0,
            "Bottom_of_range": 1200000000.0
        }]
    }

    Example 4:
    "We have seen record high use of our AI cloud tool."
    Extracted as
    {
        "Entities": [
            {"text": "record high", "category": "qualitative_desc"},
            {"text": "use", "category": "kpi_name"},
            {"text": "AI cloud tool", "category": "scope"}
        ],
        "Groups": [{
            "Source": "We have seen record high use of our AI cloud tool.",
            "Entities": [
                {"text": "record high", "category": "qualitative_desc"},
                {"text": "use", "category": "kpi_name"},
                {"text": "AI cloud tool", "category": "scope"}
            ],
            "Label": "AI cloud tool use",
            "Source Value": "record high",
            "Value": null,
            "Value_NonNumeric": "record high",
            "Is_Range": false,
            "Top_of_range": null,
            "Bottom_of_range": null
        }]
    }

    ### 3. Context ###
    - **Stock Ticker:** $tickr
    - **Fiscal Period:** $fiscal_period
    - **Time of Report:** $time_of_report

    ### 4. TASK ###
    Analyze the following text and generate the JSON output.
    If no metrics are found, output the structure with empty lists.
    Output ONLY the valid JSON object and nothing else.
    ### TEXT TO ANALYZE ###
    <text> $target_text </text>
\end{lstlisting}

\begin{lstlisting}[
    basicstyle=\scriptsize\ttfamily,
    breaklines=true,
    breakatwhitespace=true,
    frame=single,
    caption={Extraction Schema Used},
    captionpos=b,
    label={ExtractionSchema}
]
{
    "name": "financial_entity_extraction",
    "strict": True,
    "schema": {
        "type": "object",
        "properties": {
            "Entities": {
                "type": "array",
                "description": "A comprehensive list of all financial entities found in the text, classified by type.",
                "items": {
                    "type": "object",
                    "properties": {
                        "text": {
                            "type": "string",
                            "description": "The exact substring extracted from the source text."
                        },
                        "category": {
                            "type": "string",
                            "description": "The classification of the entity.",
                            "enum": [
                                "kpi_name",
                                "kpi_value",
                                "qualitative_desc",
                                "scope",
                                "date",
                                "modality"
                            ]
                        }
                    },
                    "required": ["text", "category"],
                    "additionalProperties": False
                }
            },
            "Groups": {
                "type": "array",
                "description": "Logical groupings of entities that form a single financial fact.",
                "items": {
                    "type": "object",
                    "properties": {
                        "Source": {
                            "type": "string",
                            "description": "The full text span containing the fact. Must include previous sentences if context (like Scope) is needed."
                        },
                        "Entities": {
                            "type": "array",
                            "description": "The subset of entities that belong to this specific fact.",
                            "items": {
                                "type": "object",
                                "properties": {
                                    "text": {"type": "string"},
                                    "category": {"type": "string"}
                                },
                                "required": ["text", "category"],
                                "additionalProperties": False
                            }
                        },
                        "Source Value": {
                            "type": "string",
                            "description": "The raw string representation of the value (e.g., '$10-12 million')."
                        },
                        "Label": {
                            "type": "string",
                            "description": "Unique ID. Strict Order: [Scope] [Modality] [KPI Name] [Date]. Use ONLY entities present in this group."
                        },
                        "Value": {
                            "type": ["number", "null"],
                            "description": "The numeric representation. If a range, this is the average."
                        },
                        "Value_NonNumeric": {
                            "type": ["string", "null"],
                            "description": "The qualitative description if no number exists (e.g. 'record high')."
                        },
                        "Is_Range": {
                            "type": "boolean",
                            "description": "True if the source mentions a lower and upper bound."
                        },
                        "Top_of_range": {
                            "type": ["number", "null"],
                            "description": "The upper bound of the range."
                        },
                        "Bottom_of_range": {
                            "type": ["number", "null"],
                            "description": "The lower bound of the range."
                        }
                    },
                    "required": [
                        "Source", "Entities", "Source Value", "Label",
                        "Value", "Value_NonNumeric",
                        "Is_Range", "Top_of_range", "Bottom_of_range"
                    ],
                    "additionalProperties": False
                }
            }
        },
        "required": ["Entities", "Groups"],
        "additionalProperties": False
    }
}
\end{lstlisting}
\twocolumn

\section{LLM-as-a-Judge setup}
\label{LLM-as-a-judge}
For the LLM as a judge setup, we use DeepSeek-V3.2 \citep{deepseekai2025deepseekv32pushingfrontieropen} we access through the deepseek platform\footnote{\url{https://platform.deepseek.com/}} 
We use the following prompt for the LLM-as-a-judge setup
\begin{lstlisting}[
    basicstyle=\scriptsize\ttfamily,
    breaklines=true,
    breakatwhitespace=true,
    frame=single,
    caption={The LLM as a judge prompt used.},
    captionpos=b,
    label={Lst:oneshot}
]
You are a strict financial auditor evaluating an Information Extraction system.

TASK:
Determine if the 'Model Prediction' refers to the same financial concept as the 'Ground Truth', given the context.

CONTEXT TEXT:
"{context_text}"

SHARED VALUE: {value_str}

COMPARISON:
1. Ground Truth Label: "{gt_label}"
2. Model Prediction Label: "{pred_label}"

INSTRUCTIONS:
- If the Model Prediction is a valid synonym or a reasonable extraction of the Ground Truth concept, say YES.
- If the Model Prediction captures a DIFFERENT concept (e.g., "Gross Profit" vs "Net Profit"), say NO.

OUTPUT FORMAT:
Return ONLY a JSON object:
{{
  "reasoning": "Brief explanation of your decision",
  "is_equivalent": true or false
}}
\end{lstlisting}

\end{document}